\UseRawInputEncoding

\documentclass[10pt,twocolumn,letterpaper]{article}

\usepackage[pagenumbers]{cvpr} 

\usepackage{graphicx}
\usepackage{amsmath}
\usepackage{amssymb}
\usepackage{booktabs}
\usepackage{svg}
\usepackage{float}
\usepackage{pifont}
\usepackage{diagbox}

\usepackage[pagebackref,breaklinks,colorlinks]{hyperref}

\usepackage[capitalize]{cleveref}
\crefname{section}{Sec.}{Secs.}
\Crefname{section}{Section}{Sections}
\Crefname{table}{Table}{Tables}
\crefname{table}{Tab.}{Tabs.}

\usepackage{graphicx}
\usepackage{svg}
\usepackage{pifont}
\usepackage{hyperref}

\def\cvprPaperID{*****} 
\def\confName{CVPR}
\def\confYear{2022}

\begin{document}

\title{Jersey Number Recognition using Keyframe Identification \\ from Low-Resolution Broadcast Videos}


\author{Bavesh Balaji$^*$\\
University of Waterloo\\
Waterloo, Ontario, Canada\\
{\tt\small bbalaji@uwaterloo.ca}
\and
Jerrin Bright$^*$\\
University of Waterloo\\
Waterloo, Ontario, Canada\\
{\tt\small jerrin.bright@uwaterloo.ca}
\and
Harish Prakash\\
University of Waterloo\\
Waterloo, Ontario, Canada\\
{\tt\small harish.prakash@uwaterloo.ca}
\and
Yuhao Chen\\
University of Waterloo\\
Waterloo, Ontario, Canada\\
{\tt\small yuhao.chen1@uwaterloo.ca}
\and
David A Clausi\\
University of Waterloo\\
Waterloo, Ontario, Canada\\
{\tt\small dclausi@uwaterloo.ca}
\and
John Zelek\\
University of Waterloo\\
Waterloo, Ontario, Canada\\
{\tt\small jzelek@uwaterloo.ca}}

\maketitle

\def\thefootnote{*}\footnotetext{These authors contributed equally to this work}\def\thefootnote{\arabic{footnote}}

\begin{abstract}
Player identification is a crucial component in vision-driven soccer analytics, enabling various downstream tasks such as player assessment, in-game analysis, and broadcast production. However, automatically detecting jersey numbers from player tracklets in videos presents challenges due to motion blur, low resolution, distortions, and occlusions. Existing methods, utilizing Spatial Transformer Networks, CNNs, and Vision Transformers, have shown success in image data but struggle with real-world video data, where jersey numbers are not visible in most of the frames. Hence, identifying frames that contain the jersey number is a key sub-problem to tackle. To address these issues, we propose a robust keyframe identification module that extracts frames containing essential high-level information about the jersey number. A spatio-temporal network is then employed to model spatial and temporal context and predict the probabilities of jersey numbers in the video. Additionally, we adopt a multi-task loss function to predict the probability distribution of each digit separately. Extensive evaluations on the SoccerNet dataset demonstrate that incorporating our proposed keyframe identification module results in a significant \textbf{37.81\%}  and \textbf{37.70\%} increase in the accuracies of 2 different test sets with domain gaps. These results highlight the effectiveness and importance of our approach in tackling the challenges of automatic jersey number detection in sports videos.
\end{abstract}

\begin{figure*}
{\centering
\includegraphics[width=17cm]{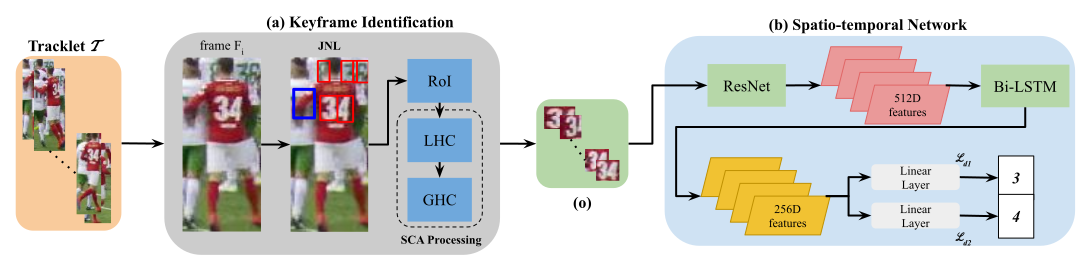}
\caption{Overview of the proposed framework for effective jersey number detection. \textbf{(a)} Keyframe identification module localizes the jersey number and eliminates outlier jersey detections. \textbf{(b)} Spatio-temporal neural network extracts the spatial and temporal context of the tracklet to identify the jersey number. (o) represents the output of a tracklet when passed to the keyframe identification module.}
\label{fig:arch}
}
\end{figure*}

\section{Introduction}

In recent years, the advent of deep learning has revolutionized various fields, enabling remarkable performance improvements. This phenomenon has now extended its influence into the realm of sports analytics, particularly in major team sports such as soccer, which enjoy extensive global viewership and participation. Teams across these sports are increasingly turning to vision-driven analytics to gain a competitive edge by evaluating player performance and making informed assessments.

At the core of player evaluation lies the crucial component of unique player identification, representing one of the most coveted research challenges in this domain. Traditionally, jersey numbers have been relied upon to establish player identification on the field. However, the fast-paced nature of the game introduces inherent challenges such as motion blur and occlusion, which hinder accurate identification. Moreover, the limited visibility of jersey numbers, typically located on the back of the player's jersey, further complicates the identification process.

Existing methods \cite{STN, vats2021multitask, Liu2019PoseGuidedRF} mainly focus on capturing spatial context and work on static images. These approaches tend to perform poorly on videos since the visibility of jersey numbers is minimal across frames. Recent works such as \cite{vats-trans, chan} try to overcome this issue by capturing temporal features using Vision Transformers and LSTMs. However, these methods still tend to give sub-optimal results on real-world tracklet data. Further investigation reveals that even with the inclusion of a temporal module, the absence of jersey numbers in the majority of frames leads to the extraction of spurious features. Consequently, the identification of keyframes, those instances capturing critical moments in the game, emerges as a vital sub-problem that demands effective solutions to ensure reliable player identification.

To address and solve the above issues, we introduce a robust Keyframe Identification (KfId) module to extract frames containing high-level features for effective jersey number recognition. The extracted frames are fed to a spatio-temporal network to model the structural and temporal context of the frames of a tracklet. We also enhance the training strategy of our model by adopting a multi-task loss function to classify each digit separately. Incorporating the KfId module in our spatio-temporal network results in a 38\% increase in test accuracy. The following items summarize the contributions of this paper:

\begin{enumerate}
    \item We propose a \textbf{\textit{keyframe identification module}} that is robust to blur and occlusions using RoI and Spatial Context Aware filtering to facilitate effective jersey number recognition.
    \item We conduct an extensive study to determine the \textbf{\textit{best training strategy for our model by experimenting with different heads for the loss function}}. We further show that digit-wise classification is the best training strategy for our model.
    \item We show that \textbf{\textit{our}} \textbf{\textit{method outperforms previous jersey number recognition methods}} on Soccernet, the largest open-source dataset collected from soccer broadcast videos for unique player identification.
\end{enumerate}

The remainder of the work is structured as follows: Sections \ref{works} discusses the related works on sports analytics for vision and jersey number identification approaches. Section \ref{method} discusses the proposed framework extensively followed by experimentation in Section \ref{exp}. Finally, the paper is concluded in Section \ref{conclusion} with directions for future work.

\section{Related Work} \label{works}





\textit{\textbf{Player identification from facial features}}
Prior to the advancements of deep learning, many works on player identification were focused on using hand-crafted facial features to recognize players. \cite{Mahmood2014AutomaticPD} use face detectors to detect faces and then perform face recognition using a database of players' faces. \cite{Ballan2007SoccerPI} use SIFT to extract facial features and localize keypoints. The extracted keypoints are then matched using a similarity measure to label players. Other works such as \cite{Senocak2018PartBasedPI, Lu2011IdentifyingPI, Lu2013LearningTT} recognize the features of the player as a whole instead of focusing on specific parts of a player. 

The caveat with these approaches is that we cannot deterministically predict the player's faces throughout a Tracklet with high confidence, since in football, broadcast cameras are usually panned at different scales at different times, making facial features negligibly available. This is especially difficult when the data parsed from such videos contains high noise due to motion blurs and occlusions. To tackle this problem several works use the jersey number to uniquely identify players.

\textit{\textbf{Jersey Number recognition from static images}}.
Gerke \textit{et al.} \cite{Gerke2015SoccerJN} recognize jersey numbers from soccer images using Convolutional Neural Networks (CNN). Li \textit{et al.} \cite{STN} use a CNN to classify the digits on a player's jersey, and mitigates the use of an extra object detection module and localizes the digits of all the players in a particular frame by using Spatial Transformer Networks (STN). Liu \textit{et al.} \cite{Liu2019PoseGuidedRF} propose a pose-guided multitask Recurrent CNN (RCNN) to jointly detect humans, human pose keypoints, and jersey numbers. Vats \textit{et al.} \cite{vats2021multitask} use a multi-task learning to recognize the digits separately and the jersey number as a whole. Bhargavi \textit{et al.} \cite{bhargavi2022knock} present a multi-stage network that takes advantage of pose to localize jersey numbers before detecting them using a secondary classifier.

\begin{figure*}
{\centering
\includegraphics[width=15cm]{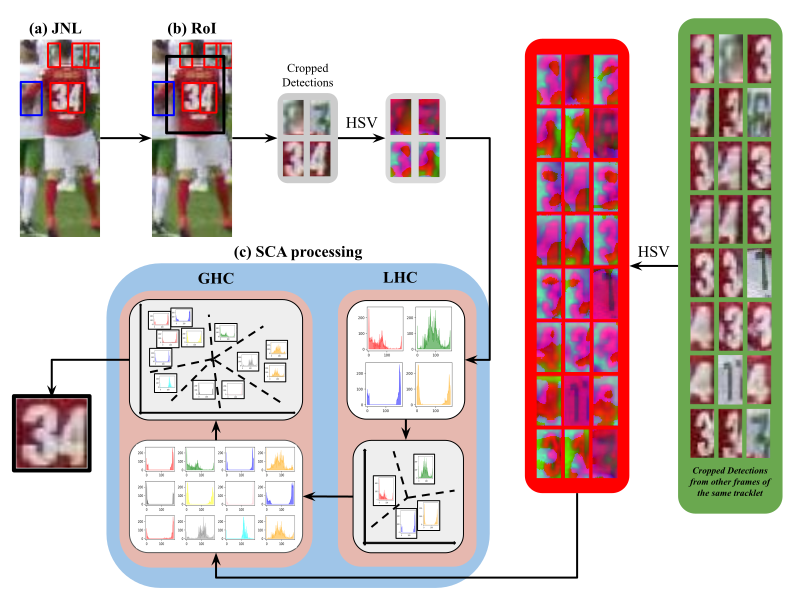}
\caption{Detailed illustration of the keyframe identification module which encompasses the following components- Jersey Number Localization (JNL), Region of Interest (RoI), Local Histogram Correlation (LHC), and Global Histogram Correlation (GHC).}
\label{fig:keyframe}
}
\end{figure*}

\noindent \textit{\textbf{Jersey Number recognition from player tracklets}}
Vats \textit{et al.} \cite{vats-trans} developed a transformer-based architecture to recognize jersey numbers from ice-hockey player tracklets. Liu \textit{et al.} \cite{liu2022deep} propose an end-to-end framework that detects players and performs unique identification through jersey number recognition from American football videos using a multi-stage approach. Chan \textit{et al.} \cite{chan} utilize LSTMs to extract temporal characteristics from player tracklets and recognize the jersey numbers of players in addition to the features extracted from a ResNet \cite{Resnet} model. Furthermore, they also employ 1D CNNs as a late score-level fusion method for classification.

All of the above methods formulate this problem as a classification problem without taking note of the inherent bias (absence of jersey numbers in many frames) in real-world data. Our method tackles the above issue by using the KfId module to identify the frames that contain useful features of the jersey number. The proposed system can help tolerate different camera Field-of-Views (FoV) during broadcast, due to varied panning and angles, and help determine player identities in a more reliable way.

\section{Methodology} \label{method}

\subsection{Keyframe Identification}

The KfID module is the key novelty of our work, which helps detect and aggregate jersey number features based on their visibility, providing a \emph{spatial-context} to the classification task. For a given player tracklet $\mathcal{T} = \{F_i : F_i \in \mathbb{R}^{H \times W \times3}\}_{i=1}^t$ consisting of $t$ frames, $KfID(\mathcal{T})  = \mathcal{T} \setminus \{F_{n_1}, F_{n_2}, ..., F_{n_k}\}$ , where $F_{n_1}, F_{n_2}, ..., F_{n_k}$ are noisy frames with diminutive digit features. In a way, our KfID module works as a selective filter, eliminating those frames that biases our Spatio-Temporal Network towards inaccurate predictions due to inconsequential features. 

The pipeline for our KfID module is as follows: For each frame $F_i$ of a given tracklet $\mathcal{T}$, the JNL module localizes all the digits in $F_i$, which are denoted as $det_i$, as shown in Equation \eqref{eq:JNL}. The detections within $F_i$ are then locally filtered ($f_{local}^{(i)}$) using RoI and Local Histogram Correlation (LHC) modules as shown in Equation \eqref{eq:fil}. The Global Histogram Correlation (GHC)  module captures the spatial similarity of the detections across frames of $\mathcal{T}$ and further filters them to get the jersey numbers of our player of interest ($f_{global}$), as shown in Equation \eqref{eq:sca}. 
\begin{equation}
    det(\mathcal{T}) = \{det_i\}_{i=1}^t = \{JNL(F_i)\}_{i=1}^t. \label{eq:JNL}
\end{equation}
\begin{equation}
   f_{local}(\mathcal{T}) = \{f_{local}^{(i)}\}_{i=1}^t = \{LHC(RoI(det_i))\}_{i=1}^t \label{eq:fil}
\end{equation}
\begin{equation}
    f_{global}(\mathcal{T}) = GHC(f_{local}(\mathcal{T})) \label{eq:sca}
\end{equation}
All the modules mentioned above will be discussed in detail in the following subsections. A detailed illustration of our $KfID$ module is shown in Figure \ref{fig:keyframe}.

\begin{figure}[H]
{\centering
\includegraphics[width=8cm]{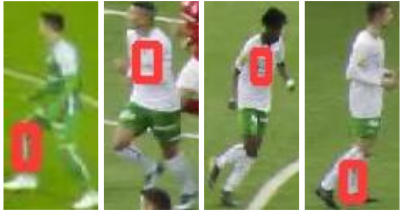}
\caption{Outlier detections from JNL module}
\label{fig:JNL}
}
\end{figure}

\subsubsection{Jersey Number Localization (JNL)}

The accuracy of jersey number classification is highly dependent on the quality of the input frames. When the input frames contain noise or other distortions, extracting useful information becomes challenging, making it difficult to accurately identify the jersey number of the target player. Thus, the presence of digits in each frame of the tracklet was first localized using an off-the-shelf detector \cite{yolov5}, fine-tuned on our dataset \cite{soccernet} to reduce the search space for the identification network. This enables narrowing down the focus to the specific regions of the frames containing the numbers.

\subsubsection{RoI-based Filtering}

The JNL module is susceptible to multiple spurious detections, as shown in Figure \ref{fig:JNL}, where a player's leg is incorrectly detected for the presence of a jersey number. To mitigate the impact of these outliers, a Region of Interest (RoI) module is incorporated since for a given Tracklet $T_i$, the jersey numbers are always localized in a specific region within a RoI. 

The RoI module is designed to filter out the JNL module's outlier detections and refine the accuracy of the predictions. Analogous to the IoU, for our specific case, we propose a custom intersection metric ($I^*$) is employed. The $I^*$ metric, defined by the following equation \eqref{eq:iou}, ensures the elimination of outlier digit detections originating from the JNL module:
\begin{equation}
    I^* = \frac{A(R_1 \cap R_2)}{min(A(R_1), A(R_2)) + eps} \label{eq:iou}
\end{equation}
where the custom intersection metric ($I^*$) is calculated based on the area of the intersection between the RoI ($R_1$) and the detection ($R_2$), as well as the areas of the RoI and the detection, respectively. Here, $A(.)$ denotes the area of a region, and $eps$ is a small constant ($10^{-7}$) added to make sure that the denominator is not $0$. The $I^*$ value is then compared to a threshold to determine the validity of the detection.



\begin{figure}[H]
{\centering
\includegraphics[width=8cm]{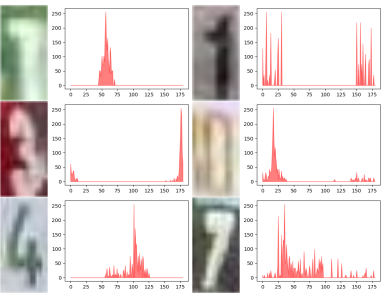}
\caption{Histogram representation of different jersey's spatial color layout.}
\label{fig:hist}
}
\end{figure}

\subsubsection{Spatial Context-Aware Processing}

\begin{figure*}[http]
{\centering
\includegraphics[width=17cm]{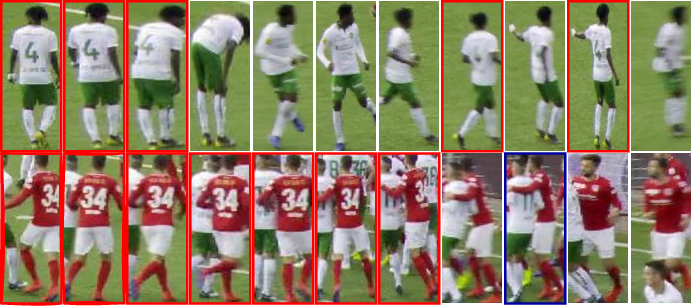}
\caption{Example frames of two different tracklets. The frames highlighted in red denote the keyframes extracted from our KfId module. Though a digit is visible in the frame highlighted in blue, our identification module ignored it, since it doesn't spatially correspond to the player of interest.} \label{fig:dataset}
}
\end{figure*}

Though the ROI module limits the search space for the detection of numbers in the frame, the visibility of the opposition team player's jersey number affects the performance of the prediction network. This problem can be predominantly seen in the second sequence of Figure \ref{fig:dataset} where the jersey number of the target player (in red jersey) is present along with the jersey number of the opposition player (in white jersey). Also, since the JNL module localizes each digit separately, a holistic representation of the jersey number is lacking. This can be seen in Figure \ref{fig:keyframe} where the JNL module captures each digit in the frame separately. Thus, in order to address both these challenges associated with the visibility and holistic representation of jersey numbers, we propose a two-stage spatial processing module comprising the LHC and GHC.

To begin with, each frame containing the detected digits undergoes conversion to the Hue, Saturation, and Value (HSV) color space. Unlike in the RGB representation, this transformation allows us to separate the color information of each component from the intensity of light, thus enabling us to better distinguish colors in various disruptive conditions. Specifically, we isolate the hue component and construct histograms to analyze the spatial correlation between different hue distributions. Figure \ref{fig:hist} depicts the contrasting color layouts for different jersey numbers. Hence, by examining the dominant color space in each frame, we gain valuable insights into the characteristics of the jersey numbers.

Next, the \textbf{LHC module} is employed to obtain a holistic representation of the jersey number by comparing correlation scores within each frame of the tracklet. If two detections in the frame are in close proximity and demonstrate similar spatial layouts (as indicated by correlation scores), they are likely to correspond to two digits forming part of the same jersey number. Consequently, these detections are merged to create a holistic representation.

Following the LHC module, we introduce the \textbf{GHC module} to specifically address the issue of jersey numbers present on opposition player jerseys. We cluster the constructed histograms of all the filtered detections of a tracklet to find out spatially similar detections. The cluster with the most number of detections is chosen and the detections in that cluster are passed on to the spatiotemporal network. By clustering the histograms based on distribution frequency within detections across all frames in a tracklet, we effectively model the spatial layout of player tracklets and distinguish the jersey number of the target player while filtering out unwanted detections within each tracklet.

\subsection{SpatioTemporal Jersey Number Recognition}

Extracting important spatial and temporal features from the pre-processed tracklets is essential for jersey number recognition. Since we have a large number of keyframes in certain tracklets, using all of the keyframes in a tracklet might lead to memory constraints and extraction of redundant features. We mitigate the above challenges by using a fixed sequence length for every tracklet and randomly sample frames while ensuring that any 2 frames sampled are at least \textit{$d$} frames apart from each other. 


Following the frame sampling process, the randomly chosen frames are passed through a pretrained ResNet-18 network pretrained on the ImageNet dataset. This step allows us to extract 512-dimensional spatial features denoted as $\mathcal{F}$. By leveraging the power of ResNet-18, we can effectively capture and encode spatial information relevant to jersey number recognition.

To capture the temporal cues within the tracklet, we further process the extracted spatial features. The spatial features $\mathcal{F}_s \in \mathbb{R}^{512}$ are fed into a bidirectional Long Short-Term Memory (bi-LSTM) network. The bi-LSTM network enables us to model both forward and backward temporal dependencies, effectively capturing the temporal dynamics present in the tracklet. As a result, we obtain $\mathcal{F}_t \in \mathbb{R}^{256}$ temporal features that encompass the temporal characteristics necessary for accurate jersey number classification.

The final step involves utilizing the obtained 256-dimensional temporal features to identify players effectively. In order to enhance our training strategy, we leverage a multi-task loss function to effectively label jersey numbers, as shown in Equation \eqref{eq:mtl}, where $d_1 \in \mathbb{R}^{11}$ and $d_2 \in \mathbb{R}^{11}$ are ground-truth digits of the jersey number, and $p_1 \in \mathbb{R}^{11}$ and $p_2 \in \mathbb{R}^{11}$ are the predictions made by the spatiotemporal network.

\begin{equation}
    L_{tot} = 0.5*L_{d1}+0.5*L_{d2} \label{eq:mtl}
\end{equation}
where,
\begin{equation}
    L_{d_1} = -\Sigma_{i=0}^{10} d_1^ilogp_1^i
\end{equation}
and
\begin{equation}
    L_{d_2} = -\Sigma_{j=0}^{10} d_2^jlogp_2^j
\end{equation}

We conducted extensive studies to ascertain the best training scheme as demonstrated in Table \ref{tab:heads}. By leveraging the spatial features extracted from the pretrained ResNet-18 and the temporal cues captured by the bidirectional LSTM network, we enable robust and accurate jersey number recognition. The overall architecture of the spatiotemporal network is illustrated in Figure \ref{fig:arch}.





\section{Experiments} \label{exp}

\subsection{Datasets} \label{dataset}

The dataset utilized in this research, referred to as "Soccernet" \cite{soccernet}, comprises a total of 4,064 player tracklets. Each tracklet is associated with a single jersey number label. To facilitate model evaluation and training, the dataset has been partitioned into four distinct subsets: training, validation, testing, and challenge sets as outlined in Table \ref{tab:data-split}. Here, the test and challenge sets are two different sets with certain domain gaps provided for evaluating the generalizability of the model. The table also shows the total number of frames extracted after the $KfID$ module. A significant drop of 87.65\% in the number of frames after KfId module shows that a major chunk of the images in each tracklet does not have a visible jersey number. 

\begin{table}[H]
  \caption{Dataset split for training, validation, and testing}
  \label{tab:data-split}
  \begin{tabular}{cccc}
    \toprule
    Dataset & Tracklets & Number of Images & Keyframes  \\
    \midrule
    Train & 1,141 & 587,543 & 68,881\\
    Validation & 286 & 146,886 & 17,220\\
    Test & 1,211 & 565,758 & 68,745\\
    Challenge & 1,426 & 750,092 & 98,504\\
    \midrule 
    Total & 4,064 & 2,052,306 & 253,350 \\
    \bottomrule
  \end{tabular}
\end{table}

Some example frames of a few tracklets can be observed in Figure \ref{fig:dataset}. The existing datasets \cite{Gerke2015SoccerJN, Liu2019PoseGuidedRF, vats2021multitask, STN} used for jersey number detection predominantly use static images, limiting modeling of temporal context, since it is essential to address problems in regards to its visibility. The different datasets available in the literature for jersey number identification are compared in Table \ref{tab:dataset_size}.

\begin{table}[H]
\centering
  \caption{Comparison of datasets in literature. $(\dagger)$ - Uses temporal data}
  \label{tab:dataset_size}
  \begin{tabular}{cc}
    \toprule
    Dataset & Number of Images\\
    \midrule
    Gerke et al \cite{Gerke2015SoccerJN} & 8,281  \\
    Liu et al \cite{Liu2019PoseGuidedRF} & 3,567  \\
    Kanav et al \cite{vats2021multitask} & 54,251  \\
    Li et al \cite{STN} & 215, 036 \\
    Kanav et al \cite{vats-trans} ($\dagger$) &  670,410 \\
    \textbf{SoccerNet} ($\dagger$) & 2,052,306\\
    \bottomrule
  \end{tabular}
\end{table}

\subsection{Implementation Details}

We adopt a ResNet18 \cite{Resnet} backbone for the spatial network with an input size of 150 $\times$ 120. Applying standard data augmentations such as ColorJitter, HorizontalFlip and normalization decreases the performance of our model. This is mainly because these augmentations decrease the quality of the data in hand, while some of them change the jersey number itself. Hence, we convert all the images to grayscale to enhance the contrast of the jersey number and the background without reducing the quality of the image.

For the RoI module, we manually preset the RoI for every image based on its dimensions. More specifically, the top left and bottom right coordinates of the RoI are ($\frac{w}{4}$, $\frac{h}{5}$) and ($\frac{3w}{4}$, $\frac{h}{2}$) respectively, where $w$ and $h$ are the width and height of the image.

For the temporal model, we evaluate and compare 3 different networks:- ViT \cite{Dosovitskiy2020AnII}, TCN \cite{Lea2016TemporalCN} and LSTM \cite{Hochreiter1997LongSM}. For ViT, we used 8 heads and 2 transformer encoder layers following \cite{vats-trans} for ideal performance. For TCNs, we used 2 1D CNN blocks, one for each digit, where each block consisted of 3 1D CNN layers followed by BatchNorm \cite{batchnorm} and ReLU. For LSTMs, we used 2 LSTM blocks for classifying each digit separately. 

All the models were trained for 20,000 iterations using a batch size of 32. We used the AdamW \cite{Loshchilov2017DecoupledWD} optimizer for the ViT model with a learning rate of 1e-4, and the Adam optimizer for the other models with a learning rate of 3e-3. Furthermore, a learning rate scheduler was employed to reduce the learning rate after every 2000 iterations for the first 6000 iterations. All the experiments were conducted on 4 NVIDIA 2080Ti GPUs with 12GB RAM each. 
\subsection{Results} 

\subsubsection{Keyframe Identification Module}

The performance enhancement of different temporal networks has been experimented with and without incorporating the KfId module as shown in Table \ref{tab:preprocess}. The aim of this experimentation is to showcase the ability of our module to improve the overall identification performance of different temporal networks.

\begin{table}[H]
\centering
  \caption{Results with and without KfId Module. ($\dagger$) - with the KfId module}
  \label{tab:preprocess}
  \begin{tabular}{ccccc}
    \toprule
    Method & Test Acc & Challenge Acc\\
    \midrule
    TCN & 27.08 & 30.17  \\
    ViT & 19.90 & 23.78  \\
    \textbf{LSTM} & \textbf{30.89} & \textbf{36.07} \\
    \midrule
    TCN ($\dagger$) & 67.54 (+40.46) & 63.81 (+33.64) \\
    ViT ($\dagger$) & 58.62 (+38.72) & 65.37 (+41.59) \\
    \textbf{LSTM} ($\dagger$) & \textbf{68.53 (+37.81)} & \textbf{73.77 (+37.70)} \\
    \bottomrule
  \end{tabular}
\end{table}

The findings presented in Table 3 reveal that the incorporation of the KfId module leads to a significant enhancement in test accuracy, as evidenced by a substantial 38\% increase compared to the performance of the best-performing model without the module. This result highlights the pivotal role played by the KfId module in augmenting the overall performance of the spatiotemporal networks. By leveraging this module, the model gains the capability to identify crucial keyframes within video sequences, capturing salient information and notable temporal changes. The discerned keyframes provide valuable contextual cues, facilitating more accurate predictions and classifications. Consequently, the observed boost in test accuracy underscores the efficacy of integrating the KfId module and suggests its potential applicability for similar models.




\subsubsection{Comparison to Existing Jersey Number Identification Works} \label{sota}

In order to assess the efficacy of our proposed architecture, we conducted a series of experiments comparing its performance against state-of-the-art networks on the SoccerNet dataset \cite{soccernet} as outlined in Table \ref{tab:sota}. We compared the performance with two different splits (Test, Challenge) from the dataset to evaluate the generalizability of the networks.

\begin{table}[H]
\centering
  \caption{Quantitative comparison with the state-of-the-art methods}
  \label{tab:sota}
  \begin{tabular}{ccc}
    \toprule
    Method & Test Acc & Challenge Acc\\
    \midrule
    Gerke et al \cite{Gerke2015SoccerJN} & 32.57 & 35.79 \\
    Kanav et al \cite{vats2021multitask} & 46.73 & 49.88 \\
    Li et al \cite{STN} & 47.85 & 50.60 \\
    Kanav et al \cite{vats-trans} & 52.91 & 58.45 \\
    \textbf{Ours} & \textbf{68.53} & \textbf{73.77} \\
    \bottomrule
  \end{tabular}
\end{table}

Our model consistently outperforms its deterministic counterparts on both the test and challenge sets. These results underscore the remarkable generalizability of our model across datasets with inherent domain gaps. 

\subsection{Ablation Study} \label{ablation}

\subsubsection{Studying Different Training Strategies}

Experimentation with different heads for the loss function is done in Table \ref{tab:heads} to determine the optimal objective and its influence on the training process. We experimented with the following heads to guide the model's prediction: 1.) Holistic ($HO$) jersey number identification aiming to predict the entire number in a single shot 2.) Digit-wise ($DW$) jersey number identification aiming to predict each digit seperately 3.) Length Control ($LC$) by aiming to control the length of the prediction by utilizing the number of digits in the number.


\begin{table}[H]
\centering
  \caption{Ablation study on different heads for the loss function}
  \label{tab:heads}
  \begin{tabular}{cccc}
    \toprule
    HO & DW & LC & Test Acc \\
    \midrule
    \ding{51} & &  & 55.71 \\
    \ding{51} & \ding{51} & & 62.39 \\
    \ding{51} & \ding{51} & \ding{51} & 65.14  \\
    & \ding{51} & \ding{51} &  63.77 \\
    & \ding{51} & &  \textbf{68.53} \\
    \bottomrule
  \end{tabular}
\end{table}

From the above table \ref{tab:heads}, we can see that performing digit-wise classification leads to better results in comparison to using holistic labels. This is mainly because of the fact that there a lot of jersey numbers in the test and challenge set that are absent in the training set. Hence, using holistic jersey number labels will lead to unseen classes in the test set which results in erroneous predictions.  Additionally, the inclusion of $LC$ in the loss function poses challenges to the network's performance, particularly due to the prevalence of jersey numbers with lengths of either 1 or 2. Relying on the jersey number length for determining the final predictions yields sub-optimal performance. These findings underscore the significance of digit-wise classification and emphasize the limitations associated with holistic labels and the incorporation of length control in the context of jersey number recognition tasks.

\subsubsection{Sequence Length}

To determine the optimal sequence length for training the temporal network, we conducted various experiments with different lengths as shown in Table \ref{tab:sequence_length}. By varying the sequence length, we aimed to evaluate the model's ability to capture temporal dependencies across frames during training.

\begin{table}[H]
\centering
  \caption{Ablation study on different training sequence length}
  \label{tab:sequence_length}
  \begin{tabular}{cc}
    \toprule
    Sequence Length & Test Acc \\
    \midrule
    10 & 62.82 \\
    20 & 65.45 \\
    30 & 66.52 \\
    \textbf{40} & \textbf{68.53}  \\
    50 & 67.03 \\
    60 & 65.80 \\
    \bottomrule
  \end{tabular}
\end{table}


The impact of sequence length on the performance of our model is presented in Table \ref{tab:sequence_length}. It is evident from the results that increasing the sequence length leads to improved model performance. However, when the sequence length surpasses 40, the accuracy of the model begins to decline. This decline suggests that using a sequence length greater than 40 causes the model to extract redundant and unnecessary features, ultimately leading to overfitting to the training set. On the other hand, employing a sequence length below 30 results in underfitting as there is insufficient data available per tracklet to capture the relevant temporal features accurately. These findings emphasize the critical role of choosing an appropriate sequence length to strike a balance between capturing meaningful temporal dynamics and preventing overfitting.

\section{Conclusion and Future Works} \label{conclusion}

In this work, we introduce and implement a robust keyframe identification module to enhance jersey number recognition from player tracklets with high-motion blur and occlusion. We further adopt spatiotemporal networks along with a multi-task loss function to perform digit-wise classification. We demonstrate the efficacy of our KfId module by conducting extensive experiments and comparing our network with state-of-the-art methods. 

One possible future improvement to our proposal is to improve our spatial model to capture more significant spatial cues even in the presence of noise.

\section{Acknowledgment}
This work was supported by Stathletes through the Mitacs Accelerate Program and Natural Sciences and Engineering Research Council of Canada (NSERC).

\bibliographystyle{ieeetr}
\bibliography{egbib}

\end{document}